\documentclass[final]{cvpr}

\pdfoutput=1

\usepackage{times}
\usepackage{epsfig}

\usepackage{url}

\usepackage{graphicx}
\usepackage{comment}
\usepackage{amsmath,amssymb,amsthm}
\usepackage{bbm}
\usepackage{multirow}

\usepackage{caption}
\usepackage{booktabs}
\usepackage{siunitx}
\usepackage{comment}
\usepackage{pifont}
\usepackage{makecell}
\usepackage{times}
\usepackage{placeins}
\usepackage{colortbl}
\usepackage{rotating}
\usepackage{array}
\usepackage{color}
\usepackage{xcolor}
\usepackage{verbatim}
\usepackage{units}
\usepackage{textcomp}
\usepackage{bbding}
\usepackage{mwe}
\usepackage{wrapfig}
\usepackage{subcaption}

\usepackage{listings}
\usepackage{bbm}
\usepackage{pifont}

\usepackage{xr}

\PassOptionsToPackage{normalem}{ulem}
\usepackage{ulem}

\usepackage[pagebackref=true,breaklinks=true,colorlinks,bookmarks=false]{hyperref}
\usepackage{cleveref}

\def\cvprPaperID{****} 
\def\confYear{CVPR 2021}

\definecolor{lightgray}{gray}{0.4}
\definecolor{verylightgray}{gray}{0.7}
\definecolor{veryverylightgray}{gray}{0.9}

\definecolor{darkgreen}{rgb}{0, 0.4, 0}
\definecolor{darkred}{rgb}{0.7, 0, 0}

\makeatletter
\renewcommand{\paragraph}{%
	\@startsection{paragraph}{4}%
	{\z@}{0.ex plus 0.5ex minus .ex}{-0.5em}{\normalsize\bf}}

\let\originalparagraph\paragraph 
\renewcommand{\paragraph}[2][.]{\originalparagraph{#2#1}}

\makeatother

\usepackage{fancyhdr}
\fancypagestyle{foot}{\fancyhf{}\fancyfoot[L]{\small{An extended version of this submission is available at \url{https://arxiv.org/abs/2103.13389}.}}}

\begin{document}

\title{Learning to Generate Novel Scene Compositions from Single Images and Videos}

\author{Vadim Sushko\\
	\small Bosch Center for Artificial Intelligence \\
	{\tt \small vadim.sushko@bosch.com}
	\and
	J{\"u}rgen Gall\\
	\small University of Bonn\\
	{\tt \small gall@iai.uni-bonn.de}
	\and
	Anna Khoreva\\
	\small Bosch Center for Artificial Intelligence\\
	{\tt \small anna.khoreva@bosch.com}
}

\makeatletter
\apptocmd\@maketitle{{\myfigure{}\par}}{}{}
\makeatother

\newcommand\myfigure{%
\begin{centering}
	\setlength{\tabcolsep}{0.1em}
	\renewcommand{\arraystretch}{0}
	\par\end{centering}
\begin{centering}
	\vspace{-1.5em}
	\hfill{}%
	\begin{tabular}{@{\hskip -0.02in}c@{\hskip 0.05in}c@{\hskip 0.05in}c@{\hskip 0.05in}c@{\hskip 0.05in}c@{\hskip 0.05in}c}
		
\textbf{Training video \& Generated samples from a single video} \tabularnewline
\includegraphics[width=0.48\linewidth]{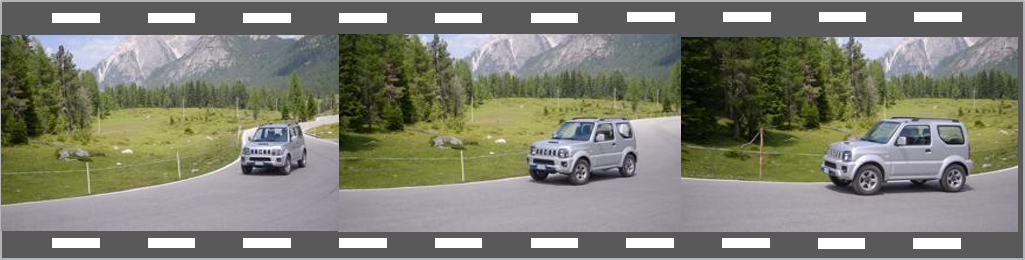} ~~
\includegraphics[width=0.48\linewidth]{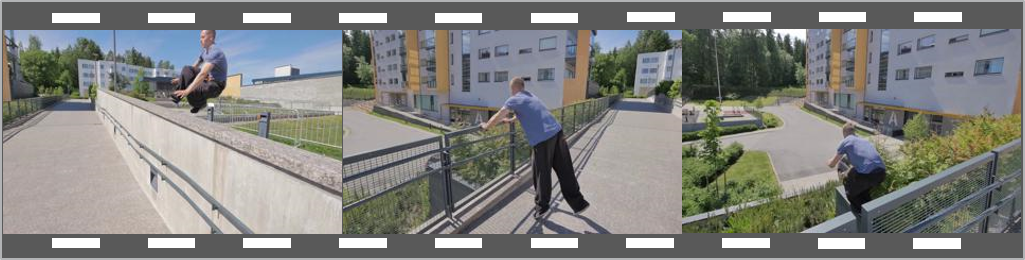}
\tabularnewline
\hspace{-0.2em}
\includegraphics[width=0.157\linewidth]{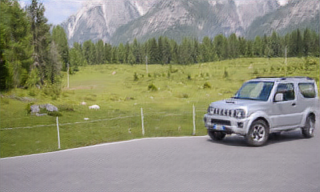} 
\includegraphics[width=0.157\linewidth]{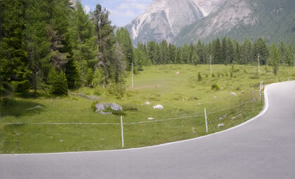}

\includegraphics[width=0.157\linewidth]{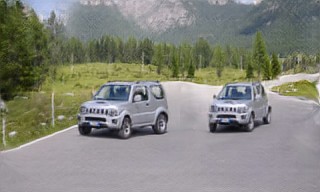}  ~~
\includegraphics[width=0.157\linewidth]{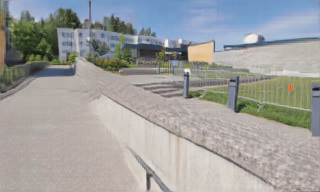}
\includegraphics[width=0.157\linewidth]{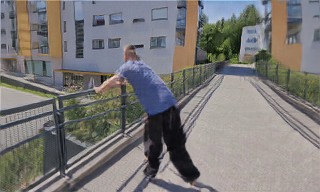}
\includegraphics[width=0.157\linewidth]{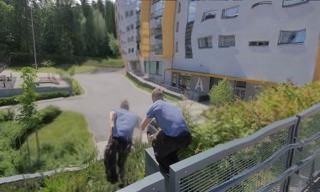}
\tabularnewline
\textbf{Training image \& Generated samples from a single image}\tabularnewline

\includegraphics[width=0.16\linewidth, height=0.08\textheight]{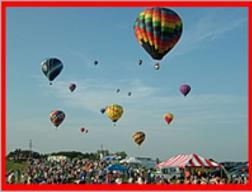}
\includegraphics[width=0.16\linewidth, height=0.08\textheight]{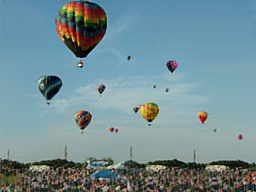}
\includegraphics[width=0.16\linewidth, height=0.08\textheight]{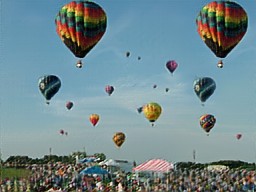}
\includegraphics[width=0.16\linewidth, height=0.08\textheight]{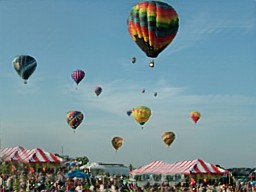}
\includegraphics[width=0.16\linewidth, height=0.08\textheight]{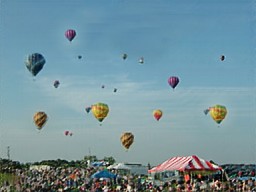}
\includegraphics[width=0.16\linewidth, height=0.08\textheight]{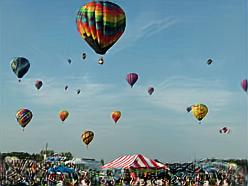}

\tabularnewline

\includegraphics[width=0.16\linewidth, height=0.08\textheight]{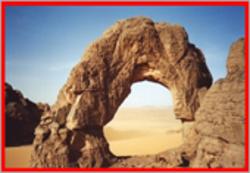}
\includegraphics[width=0.16\linewidth, height=0.08\textheight]{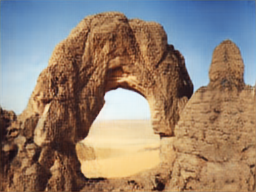} 
\includegraphics[width=0.16\linewidth, height=0.08\textheight]{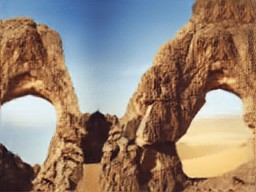}
\includegraphics[width=0.16\linewidth, height=0.08\textheight]{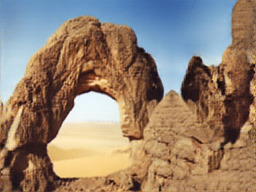}
\includegraphics[width=0.16\linewidth, height=0.08\textheight]{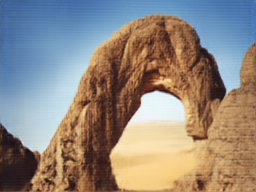} 
\includegraphics[width=0.16\linewidth, height=0.08\textheight]{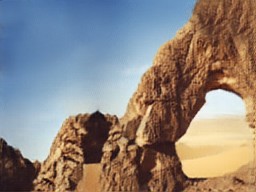} 

\tabularnewline 


	\end{tabular}
	\par\end{centering}
\vspace{-1.3ex}
\captionof{figure}{\label{fig:teaser}  Our proposed One-Shot GAN needs only one video (first two rows) or one image (last two rows) for training. At inference phase, it generates novel scene compositions with varying content and layouts. E.g., from a single video with a car on a road, One-Shot GAN can generate the scene without the car or with two cars, and for a single air balloon image, it produces layouts with different number and placement of the balloons.  
	(Original samples are shown in grey or red frames.) }
\vspace{1.5ex}
}

\maketitle
\thispagestyle{foot}

\begin{abstract}

Training GANs in low-data regimes remains a challenge, as overfitting often leads to memorization or training divergence. In this work, we introduce One-Shot GAN that can learn to generate samples from a training set as little as one image or one video. 
We propose a two-branch discriminator, with content and layout branches designed to judge the internal content separately from the scene layout realism. 
This allows synthesis of visually plausible, novel compositions of a scene, with varying content and layout, while preserving the context of the original sample.
Compared to previous single-image GAN models, One-Shot GAN achieves higher diversity and quality of synthesis. 
It is also not restricted to the single image setting, successfully learning in the introduced setting of a single video.

\end{abstract}

\section{Introduction}
\label{sec:introduction}


\begin{figure*}[t]
\begin{centering}
\setlength{\tabcolsep}{0.1em}
\renewcommand{\arraystretch}{0}
\par\end{centering}
\begin{centering}
\vspace{-1em}
\hfill{}%
\begin{tabular}{c}

\includegraphics[width=\linewidth]{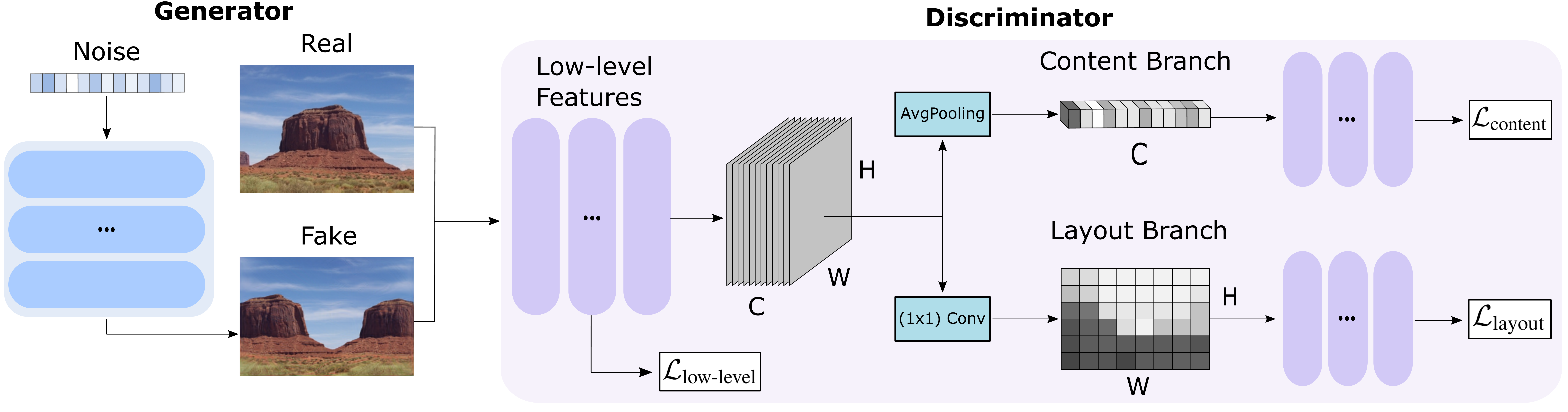} \tabularnewline

\end{tabular}\hfill{}
\par\end{centering}
\vspace{-1em}
\caption{\label{fig:model_over} One-Shot GAN. The two-branch discriminator judges the content distribution separately from the scene layout realism and thus enables the generator to produce images with varying content and global layouts. See Sec.~\ref{sec:method_discriminator} for details.
}
\vspace{-0.5em}
\end{figure*}

Without sufficient training data, state-of-the-art GAN \cite{karras2019analyzing, Brock2019, schnfeld2021you} models are prone to overfitting, which often leads to mode collapse and training instabilities \cite{Shaham2019SinGANLA,Hinz2020ImprovedTF}.
This dependency on availability of training data limits the applicability of GANs 
in domains where collecting a large dataset is not feasible. In some real-world applications, collection even of a small dataset remains challenging.
 It may happen that rare objects or events are present only in one image or in one video, and it is difficult to obtain a second one. This, for example, includes pictures of exclusive artworks or videos of traffic accidents recorded in extreme conditions. Enabling learning of GANs in such \textit{one-shot} scenarios \cite{sushko2021one} has thus a potential to improve their utilization in practice.
Previous work \cite{Shaham2019SinGANLA,Hinz2020ImprovedTF} studied one-shot image generation in the context of learning from a \textit{single image}. In this work, we introduce a novel setup of learning to generate new images from frames of a \textit{single video}. In practice, recoding a video lasting for several seconds can take almost as little effort as collecting one image. However, a video contains much more information about the scene and the objects of interest (e.g., different poses and locations of objects, various camera views). Learning from a video can enable generation of images of higher quality and diversity, while still operating in a one-shot mode, and therefore can improve its usability for applications. 

\begin{figure*}[h!]
\begin{centering}
\setlength{\tabcolsep}{0.0em}
\renewcommand{\arraystretch}{0}
\par\end{centering}
\begin{centering}
\vspace{-1.5em}
\begin{tabular}{@{\hskip -0.03in}c@{\hskip 0.0in}c@{\hskip 0.05in}c@{\hskip 0.11in}c@{\hskip 0.05in}c@{\hskip 0.11in}c@{\hskip 0.05in}c@{}}
& \multicolumn{2}{c}{SinGAN~\cite{Shaham2019SinGANLA}} &\multicolumn{2}{c}{FastGAN~\cite{anonymous2021towards}}  & \multicolumn{2}{c}{One-Shot GAN}  
\tabularnewline	

\multirow{-2}{*}{\begin{tabular}{c}  Training image \\\includegraphics[width=0.13\linewidth, height=0.07\textheight]{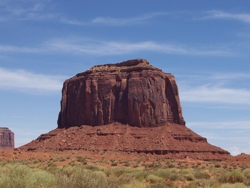} \end{tabular}} &
\includegraphics[width=0.13\linewidth, height=0.07\textheight]{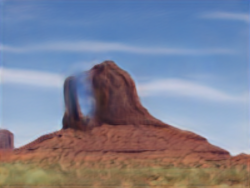} & 
\includegraphics[width=0.13\linewidth, height=0.07\textheight]{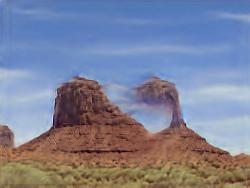} & 
\includegraphics[width=0.13\linewidth, height=0.07\textheight]{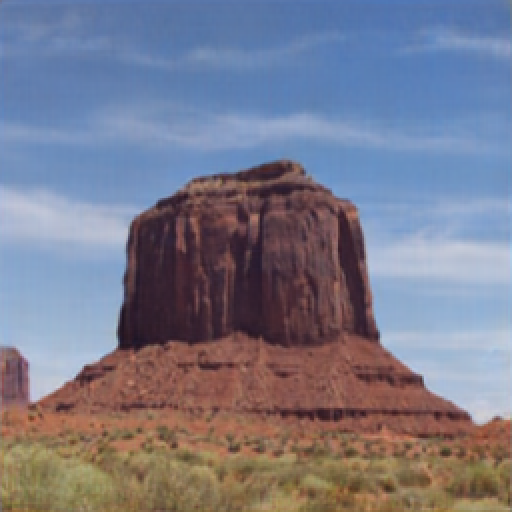} & 
\includegraphics[width=0.13\linewidth, height=0.07\textheight]{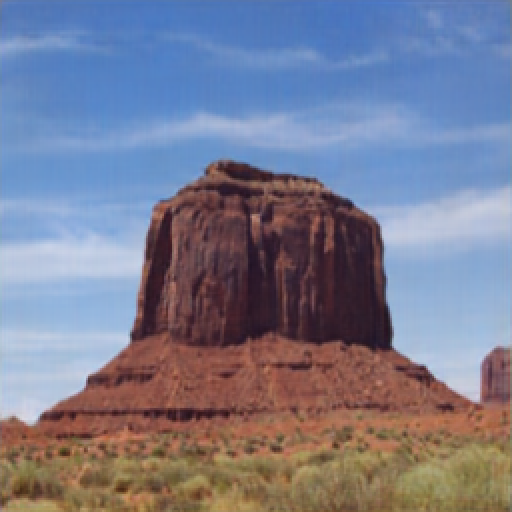} & 
\includegraphics[width=0.13\linewidth, height=0.07\textheight]{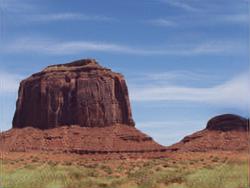} & 
\includegraphics[width=0.13\linewidth, height=0.07\textheight]{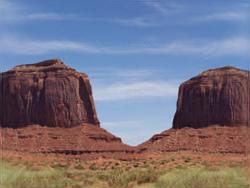} \tabularnewline 

& 
\includegraphics[width=0.13\linewidth, height=0.07\textheight]{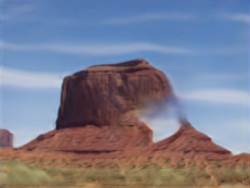} & 
\includegraphics[width=0.13\linewidth, height=0.07\textheight]{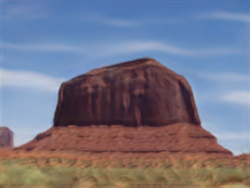} & 
\includegraphics[width=0.13\linewidth, height=0.07\textheight]{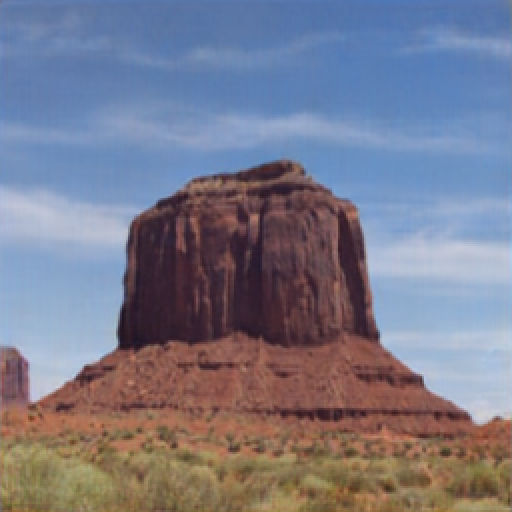} & 
\includegraphics[width=0.13\linewidth, height=0.07\textheight]{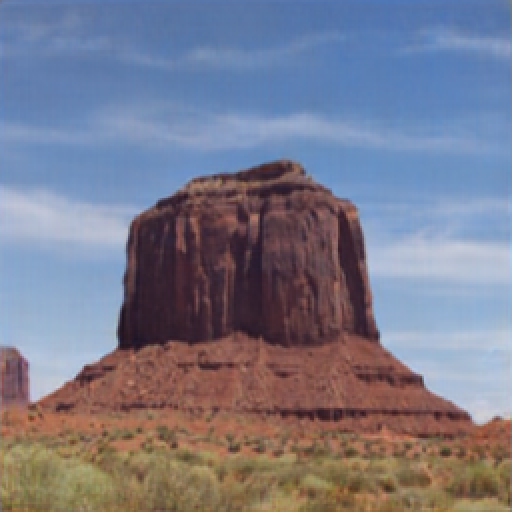} & 
\includegraphics[width=0.13\linewidth, height=0.07\textheight]{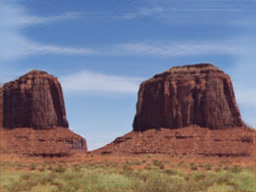} & 
\includegraphics[width=0.13\linewidth, height=0.07\textheight]{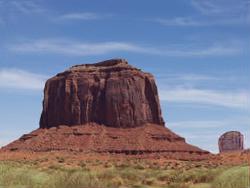}

\end{tabular}\hfill{}
\par\end{centering}
\vspace{-0.5em}
\caption{\label{fig:qual_image} Comparison with other methods in the single image setting. Single-image GAN of~\cite{Shaham2019SinGANLA} is prone to incoherently shuffle image patches (e.g. sky textures appear on the ground), and the few-shot FastGAN model~\cite{anonymous2021towards} collapses to producing the original image or its flipped version. In contrast, One-Shot GAN produces diverse images with realistic \textit{global} layouts.}
\vspace{-1em}
\end{figure*}

To mitigate overfitting in the one-shot mode, recent 
single-image GAN models~\cite{Shaham2019SinGANLA,Hinz2020ImprovedTF} proposed to learn an image patch-based distribution at different image scales. 
Though these models overcome memorization, producing different versions of a training image, they cannot learn high-level semantic properties of the scene. They thus
often suffer from incoherent shuffling of image patches, distorting objects and producing unrealistic layouts (see Fig. \ref{fig:qual_image} and \ref{fig:qual_video}). Other low-data GAN models, such as FastGAN~\cite{anonymous2021towards}, have problems with memorization when applied in the one-shot setting (see Sec. \ref{sec:experiments}).
In this work, we go beyond patch-based learning, seeking to generate novel plausible compositions of objects in the scene, while preserving the original image context.
Thus, we aim to keep novel compositions visually plausible, with objects preserving their appearance, and the scene layout looking realistic to a human eye.

To this end, we introduce One-Shot GAN, an unconditional single-stage GAN model, which can generate images that are significantly different from the original training sample while preserving its context.
This is achieved by two key ingredients: the novel design of the discriminator and the proposed diversity regularization technique for the generator. The new One-Shot GAN discriminator has two branches, responsible for judging the content distribution and the scene layout realism of images separately from each other. Disentangling the discriminator decision about the content and layout helps to prevent overfitting and provides more informative signal to the generator. To achieve high diversity of generated samples, we also extend the regularization technique of \cite{Yang2019DiversitySensitiveCG,choi2020stargan} to one-shot unconditional image synthesis. 
As we show in Sec. \ref{sec:experiments}, The proposed One-Shot GAN generates high-quality images that are significantly different from training data. One-Shot GAN is the first model that succeeds in learning from both single images and videos, improving over prior work \cite{Shaham2019SinGANLA,anonymous2021towards} in image quality and diversity in these one-shot regimes. 
\section{One-Shot GAN}
\label{sec:method}

\begin{table*}[t]

\vspace{-0em}
	\setlength{\tabcolsep}{0.2em}
	\renewcommand{\arraystretch}{0.95}
	\centering
	
	\begin{tabular}{c|c|c|c|c||c|c|c|c}
		& \multicolumn{4}{c||}{\normalsize{} Single Image } & \multicolumn{4}{c}{\normalsize{} Single Video } \tabularnewline
		\multirow{2}{*}{\normalsize{} Method } &  \multirow{2}{*}{{\small{} SIFID$\downarrow$}} & \multirow{2}{*}{\small{} LPIPS$\uparrow$ } & \multirow{2}{*}{\small{} MS-SSIM~$\downarrow$} & {\small{}Dist.~ } & \multirow{2}{*}{{\small{} SIFID$\downarrow$}} & \multirow{2}{*}{\small{} LPIPS$\uparrow$ } & \multirow{2}{*}{\small{} MS-SSIM~$\downarrow$} & {\small{}Dist.~ } \tabularnewline
		&  & & & to train & & & & to train \tabularnewline
		
		\hline 	 \hline 	
		
		{\small{} SinGAN \cite{Shaham2019SinGANLA} } & \small{} 0.13 & \small{} 0.26 & \small{} 0.69 & \small{0.30} & \color{darkred} \small{2.47}  &  \small{0.32}  & \small{0.65} &    \small{0.51} \tabularnewline
				
		{\small{} FastGAN \cite{anonymous2021towards}} & \small{} 0.13 & \small{} 0.18 & \small{} 0.77 & \color{darkred} \small{0.11} & \small{0.79} & \textbf{\small{0.43}} & \small{0.55} &  \color{darkred} \small{0.13}  \tabularnewline		
		
		{\small{} One-Shot GAN} & \small{} \textbf{0.08} & \small{} \textbf{0.33} & \small{} \textbf{0.63}  & \small{0.37} & \textbf{\small{0.55}} & \textbf{\small{0.43}} & \textbf{\small{0.54}}  &  \small{0.34}
		\end{tabular}
	\vspace{-0.5em}
	\caption{Comparison in the Single Image and Single Video settings on DAVIS-YFCC100M \cite{pont20172017,thomee2016yfcc100m} dataset.}
\label{table:comp_single_image} %
\vspace{-1em}
\end{table*}

\paragraph{Content-Layout Discriminator} \label{sec:method_discriminator}

We introduce a solution to overcome the memorization effect but still to generate images
of high quality in the one-shot setting. Building on the assumption that to produce realistic and diverse images the generator should learn the appearance of objects and combine them in a globally-coherent way in an image, we propose a discriminator judging the \textit{content} distribution of an image separately from its \textit{layout} realism. 
To achieve the disentanglement, we design a two-branch discriminator architecture, with separate content and layout branches (see Fig.\ref{fig:model_over}). Our discriminator $D$ consists of the low-level feature extractor $D_{low\text{-}level}$, the content branch $D_{content}$, and the layout branch $D_{layout}$. Note that the branching happens after an intermediate layer in order to learn a relevant representation.
$D_{content}$ judges the content of this representation irrespective from its spatial layout, while $D_{layout}$, inspects only the spatial information. Inspired by the attention modules of \cite{park2018bam, woo2018cbam}, we extract the \textit{content} from intermediate representations by aggregating spatial information via global average pooling, and obtain \textit{layout} by aggregating channels via a simple $(1\times 1)$ convolution. This way, the content branch judges the fidelity of objects composing the image independent of their spatial location, while the layout branch is sensitive only to the realism of the global scene layout. Note that $D_{content}$ and $D_{layout}$ receive only limited information from previous layers, which prevents overfitting. This helps to overcome the memorization of training data and to produce different images.

\paragraph{Diversity regularization}

\begin{figure*}[t]
\begin{centering}
\setlength{\tabcolsep}{0.1em}
\renewcommand{\arraystretch}{0}
\par\end{centering}
\begin{centering}
\vspace{-1em}
\hfill{}%
\begin{tabular}{@{\hskip -0.05in}c@{\hskip 0.05in}c@{\hskip 0.05in}c@{\hskip 0.05in}c@{\hskip 0.05in}|c@{\hskip 0.05in}c@{\hskip 0.05in}c@{\hskip 0.05in}c}

\rotatebox{90}{ \hspace{0.5ex}\begin{tabular}{c} \small Video \\ \small frames\end{tabular}} &  \multicolumn{3}{c|}{\hspace{-2.05ex} \includegraphics[width=0.463\linewidth, height=0.075\textheight]{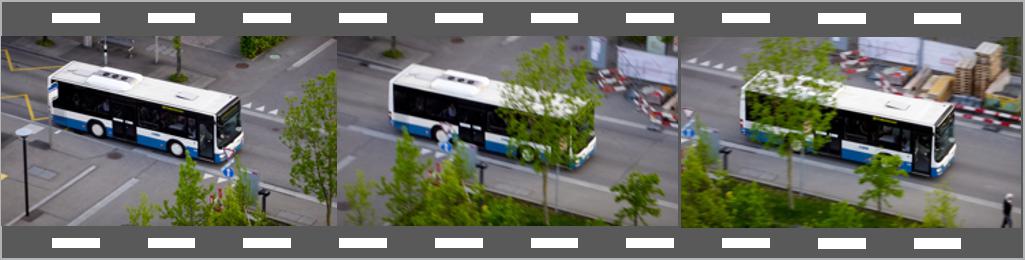} } &
\multicolumn{3}{c}{\includegraphics[width=0.463\linewidth, height=0.075\textheight]{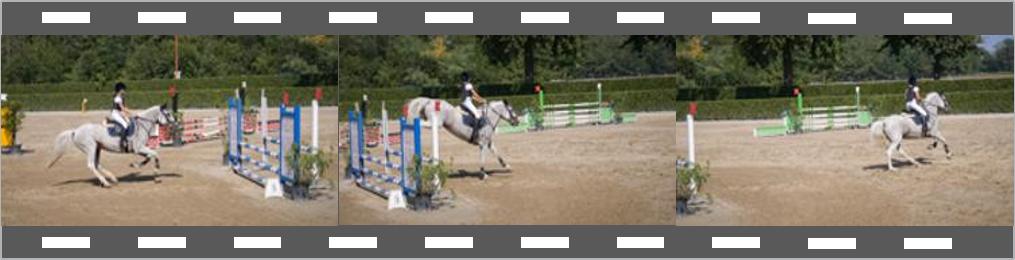}} \tabularnewline
\rotatebox{90}{\hspace{-.7ex}\begin{tabular}{c} \small SinGAN \\ \small ~\cite{Shaham2019SinGANLA}\end{tabular}}&
\includegraphics[width=0.15\linewidth, height=0.06\textheight]{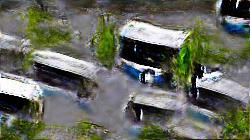} & 
\includegraphics[width=0.15\linewidth, height=0.06\textheight]{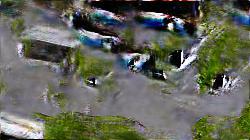} & 
\includegraphics[width=0.15\linewidth, height=0.06\textheight]{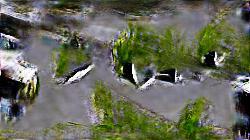} &
\includegraphics[width=0.15\linewidth, height=0.06\textheight]{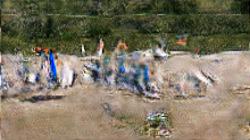} & 
\includegraphics[width=0.15\linewidth, height=0.06\textheight]{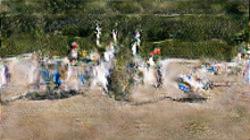} &
\includegraphics[width=0.15\linewidth, height=0.06\textheight]{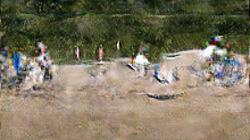}\tabularnewline
\rotatebox{90}{\hspace{-.8ex}\begin{tabular}{c} \small FastGAN \\\small ~\cite{anonymous2021towards}\end{tabular}}&
\includegraphics[width=0.15\linewidth, height=0.06\textheight]{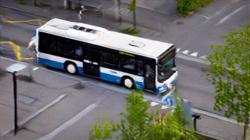} & 
\includegraphics[width=0.15\linewidth, height=0.06\textheight]{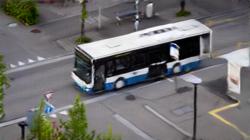} & 
\includegraphics[width=0.15\linewidth, height=0.06\textheight]{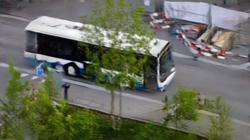} & 
\includegraphics[width=0.15\linewidth, height=0.06\textheight]{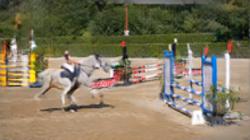} &
\includegraphics[width=0.15\linewidth, height=0.06\textheight]{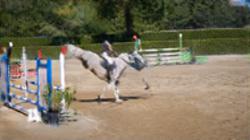} & 
\includegraphics[width=0.15\linewidth, height=0.06\textheight]{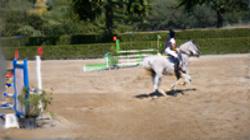}\tabularnewline
\rotatebox{90}{\hspace{-.9ex}\begin{tabular}{c} \small One-Shot \\ \small GAN\end{tabular}}&
\includegraphics[width=0.15\linewidth, height=0.06\textheight]{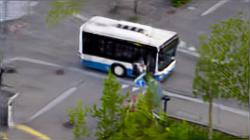} &
\includegraphics[width=0.15\linewidth, height=0.06\textheight]{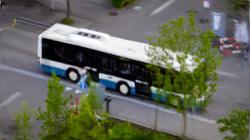} & 
\includegraphics[width=0.15\linewidth, height=0.06\textheight]{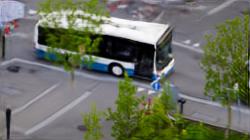} & 
\includegraphics[width=0.15\linewidth, height=0.06\textheight]{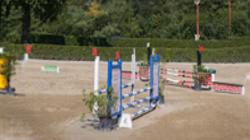} &
\includegraphics[width=0.15\linewidth, height=0.06\textheight]{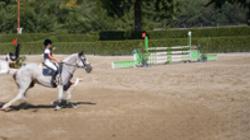} & 
\includegraphics[width=0.15\linewidth, height=0.06\textheight]{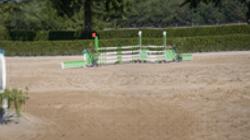}

\end{tabular}\hfill{}
\par\end{centering}
\vspace{-0.5em}
\caption{\label{fig:qual_video} Comparison with other methods in the single video setting. While other models fall into reproducing the training frames or fail to learn textures, One-Shot GAN generates high-quality images significantly different from the original video.
 }
\vspace{-1em}
\end{figure*}

To improve variability of generated samples, we propose to add diversity regularization (DR) loss term $\mathcal{L}_{DR}$ to the objective. 
Previously proposed regularization terms \cite{Yang2019DiversitySensitiveCG, choi2020stargan} aimed to encourage the generator to produce different outputs depending on the input latent code, in such a way that the generated samples with closer latent codes should look more similar to each other, and vice
versa. In contrast, in the one-shot
image synthesis setting, the perceptual distance of the generated
images should not be dependent on the distance between
their latent codes. As we operate in one semantic domain, the generator should produce images that are in-domain but more or less equally different from each other and substantially
different from the original sample. Thus, we propose to encourage the generator to produce perceptually different image samples independent of their distance in the latent space. $\mathcal{L}_{DR}$ is expressed as follows:
\begin{align}
\mathcal{L}_{DR} (G) = \mathbb{E}_{z_1,z_2}\big[\frac{1}{L}\sum_{l=1}^{L}\|G^{l}(z_1)-G^{l}(z_2)\|)\big], \label{loss_DR}
\end{align}
where $\|\cdot\|$ denotes the $L1$ norm, $G^{l}(z)$ indicates features extracted from the $l$-th block of the generator $G$ given the latent code $z$. Contrary to prior work, we compute the distance between samples in the feature space, enabling more meaningful diversity of the generated images, as different generator layers capture various image semantics, inducing both high- and low-level diversity.

\paragraph{Final objective}
We compute adversarial loss for each discriminator part: $D_{low\text{-}level}$,  $D_{content}$, and $D_{layout}$. This way, the discriminator decision is based on low-level details of images, such as textures, and high-scale properties, such as content and layout. The overall adversarial loss is
\begin{align}  \vspace{-5pt}
& \mathcal{L}_{adv} (G,D)  = \mathcal{L}_{D_{content}} + \mathcal{L}_{D_{layout}} + 2 \mathcal{L}_{D_{low\text{-}level}}, \label{eq:loss_d_2}
\end{align} 
where $\mathcal{L}_{D_{*}}$ is the binary cross-entropy $\mathbb{E}_x[\log D_{*}(x)] + \mathbb{E}_z[\log (1-D_{*}(G(z)))]$ for real image $x$ and generated image $G(z)$.
As the two branches of the discriminator operate at high-level image features, contrary to only one $D_{low\text{-}level}$ operating at low-level features, we double the weighting for the $\mathcal{L}_{D_{low\text{-}level}}$ loss term. This is done in order to properly balance the contributions of different feature scales and encourage the generation of images with good low-level details, coherent contents and layouts. 

The overall One-Shot GAN objective can be written as:
\begin{align}
\min_{G} \max_{D} \hspace{0.5em} \mathcal{L}_{adv} (G,D) - \lambda \mathcal{L}_{DR} (G), \label{loss_full}
\end{align}
where $\lambda$ controls the strength of the diversity regularization and $\mathcal{L}_{adv}$ is the adversarial loss in Eq.~\ref{eq:loss_d_2}.

\paragraph{Implementation} \label{sec: train_det}
The One-Shot GAN discriminator uses ResNet blocks, following \cite{Brock2019}. 
We use three ResNet blocks before branching and four blocks for the content and layout branches.
We employ standard image augmentation strategies for the discriminator training, following \cite{Karras2020TrainingGA}. $\lambda$ for $\mathcal{L}_{DR}$ in  Eq.~\ref{loss_full} is set to $0.15$.
We use the ADAM optimizer with $(\beta_1, \beta_2) = (0.5, 0.999)$, a batch size of $5$ and a learning rate of $0.0002$ for both $G$ and $D$.  
\section{Experiments}
\label{sec:experiments}

\paragraph{Evaluation settings} We evaluate One-Shot GAN on two different one-shot settings: training on a single image and a single video. 
We select 15 videos from DAVIS~\cite{pont20172017} and YFCC100M \cite{thomee2016yfcc100m} datasets. In the Single Video setting, we use all frames of a video as training images, while for the Single Image setup we use only one middle frame. The chosen videos last for 2-10 seconds and consist of 60-100 frames. To assess the quality of generated images, we measure single FID (SIFID) \cite{Shaham2019SinGANLA}. Image diversity is assessed by the average LPIPS \cite{Alexey2016} and MS-SSIM \cite{wang2003multiscale} across pairs of generated images. To verify that the models do not simply reproduce the training set, we report average LPIPS to the nearest image in the training set, augmented in the same way as during training (Dist. to train). We compare our model with a single image method SinGAN \cite{Shaham2019SinGANLA} and with a recent model on few-shot image synthesis, FastGAN \cite{anonymous2021towards}.

\paragraph{Main results} Table~\ref{table:comp_single_image} presents quantitative comparison between the models in the Single Image and Video settings, while the respective visual results are shown in Fig.~\ref{fig:qual_image} and \ref{fig:qual_video}. As seen from Table~\ref{table:comp_single_image}, One-Shot GAN notably outperforms other models in both quality and diversity metrics.
Importantly, our model is the only one which successfully learns from both single images and single videos. 


As seen from Fig. \ref{fig:teaser} and \ref{fig:qual_image}, in the Single Image setting, One-Shot GAN produces diverse samples of high visual quality. For example, our model can change the number of rocks on the background or change their shapes. Note that such changes keep appearance of objects, preserving original content, and maintain scene layout realism. In contrast, single-image method SinGAN disrespects layouts (e.g. sky textures may appear below horizon), and is prone to modest diversity, especially around image corners. This is reflected in higher SIFID and lower diversity in Table \ref{table:comp_single_image}. The few-shot FastGAN suffers from memorization, only reproducing the training image or its flipped version. In Table~\ref{table:comp_single_image} this is reflected in low diversity and small Dist. to train (in red). 

In the proposed Single Video setting, there is much more information to learn from, so generative models can learn more interesting combinations of objects and scenes. Fig. \ref{fig:teaser} and \ref{fig:qual_video} show images generated by the models in this setting. One-Shot GAN produces plausible images that are substantially different from the training frames, adding/removing objects and changing scene geometry.
For example, having seen a bus following a road, One-Shot GAN varies the length of a bus and placement of trees. For the video with an equestrian competition, our model can remove a horse from the scene and change the jumping obstacle configuration. In contrast, SinGAN, which is tuned to learn from a single image, does not generalize to this setting, producing "mean in class" textures and failing to learn appearance of objects (low diversity and very high SIFID). FastGAN, on the other hand, learns high-scale scene properties, but fails to augment the training set with non-trivial changes, having a very low distance to the training data ($0.13$ in Table \ref{table:comp_single_image}).

Table \ref{table:comp_single_image} confirms that the proposed two-branch discriminator in combination with diversity regularization manages to overcome the memorization effect, achieving high distance to training data in both settings ($0.37$ and $0.34$). This means that One-Shot GAN augments the training set with structural transformations that are orthogonal to standard data augmentation techniques, such as horizontal flipping or color jittering. To achieve this, the model requires as little data as one image or one short video clip. We believe, such ability can be especially useful to generate samples for augmentation of limited data, for example by creating new versions of rare examples.

\vspace{-0.5em}

%
%
\section{Conclusion}
\label{sec:conclusion}

We propose One-Shot GAN, a new unconditional generative model operating at different one-shot settings, such as learning from a single image or a single video. At such low-data regimes, our model mitigates the memorization problem and generates diverse images that are structurally different from the training set. 
Particularly, our model is capable of synthesizing images with novel views and different positions of objects, preserving their visual appearance. We believe, such structural diversity provides a useful tool for image editing applications, as well as for data augmentation in domains, where data collection remains challenging.

{\small
\bibliographystyle{ieee_fullname}
\bibliography{references}
}

\end{document}